# Review

# Unmasking bias in artificial intelligence: a systematic review of bias detection and mitigation strategies in electronic health record-based models

Feng Chen, MS*,[1,2], Liqin Wang, PhD[1,3], Julie Hong, HS[4], Jiaqi Jiang, MS[1], Li Zhou, MD, PhD[1,3]

[1]Department of Biomedical Informatics, Harvard Medical School, Boston, MA 02115, United States, [2]Department of Biomedical Informatics and Health Education, University of Washington, Seattle, WA 98105, United States, [3]Division of General Internal Medicine and Primary Care, Brigham and Women's Hospital, Boston, MA 02115, United States, [4]Wellesley High School, Wellesley, MA 02481, United States

*Corresponding author: Feng Chen, MS, University of Washington, 4751 12th Ave NE, Apt 306, Seattle, WA 98105, United States (fengc9@uw.edu)

## Abstract

**Objectives:** Leveraging artificial intelligence (AI) in conjunction with electronic health records (EHRs) holds transformative potential to improve healthcare. However, addressing bias in AI, which risks worsening healthcare disparities, cannot be overlooked. This study reviews methods to handle various biases in AI models developed using EHR data.
**Materials and Methods:** We conducted a systematic review following the Preferred Reporting Items for Systematic Reviews and Meta-analyses guidelines, analyzing articles from PubMed, Web of Science, and IEEE published between January 01, 2010 and December 17, 2023. The review identified key biases, outlined strategies for detecting and mitigating bias throughout the AI model development, and analyzed metrics for bias assessment.
**Results:** Of the 450 articles retrieved, 20 met our criteria, revealing 6 major bias types: algorithmic, confounding, implicit, measurement, selection, and temporal. The AI models were primarily developed for predictive tasks, yet none have been deployed in real-world healthcare settings. Five studies concentrated on the detection of implicit and algorithmic biases employing fairness metrics like statistical parity, equal opportunity, and predictive equity. Fifteen studies proposed strategies for mitigating biases, especially targeting implicit and selection biases. These strategies, evaluated through both performance and fairness metrics, predominantly involved data collection and preprocessing techniques like resampling and reweighting.
**Discussion:** This review highlights evolving strategies to mitigate bias in EHR-based AI models, emphasizing the urgent need for both standardized and detailed reporting of the methodologies and systematic real-world testing and evaluation. Such measures are essential for gauging models' practical impact and fostering ethical AI that ensures fairness and equity in healthcare.

## Background and significance

The rapid advancement of artificial intelligence (AI) in healthcare, particularly through the utilization of large-scale, real-world electronic health records (EHRs) data, has revolutionized medical research and clinical decision-making (CDS). Especially in the past decade, EHRs have become increasingly mature and widespread, providing a vast and rich source of data for AI development.[1] Unlike AI applications aimed at enhancing the broader healthcare ecosystem, such as public health surveillance, drug discovery, and healthcare operations,[2] AI models built from EHR data are more intimately connected with individual patient records and the CDS processes.[3] EHRs, encompassing diverse patient data, including demographics, lab results, diagnoses, and treatments, are uniquely suited for developing data-driven AI models for predictive analytics with applications ranging from risk identification and disease progression prediction,[4] and outcome forecasting.[5]

However, the integration of AI with EHR data, while crucial for advancing CDS and medical research, confronts challenges from biases inherent in EHR data and AI models. These challenges include inconsistencies in documentation, variations in data quality, and model inaccuracies.[6,7] Such inaccuracies, manifesting as analytical errors and skewed outcomes often due to disproportionate dataset representation (eg, overrepresentation or underrepresentation of certain patient demographics, health conditions, or treatment types in the training data) can lead to differential performance across patient subgroups, potentially exacerbating healthcare disparities.[8,9] Biases can emerge at various stages of the model development lifecycle,[10–12] underlining the importance of developing and implementing strategies for effective detection, mitigation, and evaluation of biases.[6,7] Tailoring these strategies to the unique characteristics of EHR data and AI models is essential for ensuring their efficient and equitable application, thereby facilitating successful AI integration in healthcare.[13]

While a few scoping reviews have been conducted to understand bias in broader medical AI,[14,15] a focused review on AI models derived from EHR data is notably absent. This gap underscores the need for a systematic review to identify, summarize, and propose strategies for managing these biases. This study aims to bridge this gap by analyzing existing research on AI biases within EHR-based models.

## Objective

This study aims to systematically review and synthesize the current literature on bias in AI models built from EHR data, focusing on the identification, evaluation, and mitigation strategies for major types of biases across the model development cycle. The goal is to enhance understanding of AI bias management within EHR-based applications and highlight research directions to reduce potential AI impacts on healthcare disparities.

## Materials and methods

### Data sources and searches

This review was conducted in compliance with the 2021 PRISMA (Preferred Reporting Items for Systematic Reviews and Meta-analyses) guidelines, as illustrated in Figure 1. We conducted systematic searches on 3 relevant publication databases (PubMed/MEDLINE, Web of Science, and the Institute of Electrical and Electronics Engineers) to retrieve articles published between January 1, 2010, and December 17, 2023. Search queries used for individual databases are available in Table S1.

### Inclusion and exclusion criteria

We included articles that: (1) were written in English; (2) contained metadata (authors, title, publication year) and full text; (3) were published between January 1, 2010 and December 17, 2023; (4) focused on EHR-based AI models (ie, EHR data were used for model training, testing, and validation); and (5) evaluated bias, clearly describing its impact on healthcare disparity and detailing bias handling approaches. Studies centered on imaging and device-related data were excluded due to the distinct nature of potential biases that arise from the format, collection, utilization, and interpretation of data between EHRs and medical images/devices.

### Article screening process

We retrieved titles and abstracts from 3 databases and removed the duplicates. Each of the remaining titles and abstracts was then screened by at least 2 reviewers (Feng Chen, Julie Hong, Jiaqi Jiang). The full-text screening was conducted by Feng Chen, Liqin Wang, and Julie Hong with each article reviewed by at least 2 reviewers independently. Any inconsistencies among reviewers were resolved through team meetings for consensus.

### Data extraction

Data was initially extracted by Feng Chen and Jiaqi Jiang and reviewed by Liqin Wang and Li Zhou. We thoroughly examined the full text of the remaining articles to extract pertinent components for our analysis. The data extraction process covered 3 key categories: bibliographic data, information related to the AI model, and specifics of bias and fairness. Bibliographic data included the title, authors, and year of publication of the study. Information related to the AI model comprised the source of EHR data, sample size, and the primary objectives and tasks of the AI models. Lastly, bias/fairness-specific information included the types of bias reported in each study, the strategies employed to detect or mitigate the bias, evaluation metrics used to measure the biases.

### Categorization of bias types

The categorization of bias types in EHR-based AI models involved a structured 2-step process. First, we examined the biases reported in the selected studies to reflect the current understanding of biases in AI research. This initial analysis provided a direct insight into the prevalent issues within the field. Second, we expanded our scope by integrating insights from the broader literature on healthcare AI, including works beyond our systematic review. Our analyses were further enhanced by consolidating established bias risk assessment tools, such as ROBINS-I,[16] ROBINS-E,[17] and PROBAST,[18] and relevant review articles, such as Mehrabi et al[19] which categorized potential biases in AI applications. By combining these methodologies, we defined and categorized the major types of bias present in EHR-based AI models, ensuring a thorough understanding of the bias landscape in this domain.

### Bias analyses workflow construction

We created a framework, illustrated in Figure 2, to categorize and examine methods for addressing bias in AI model development. This framework identifies the major potential biases (examined from the above step) that may arise during 3 key stages: data collection and preparation, model training and testing, and model deployment. For each stage, we identified specific types of bias and reviewed targeted mitigation strategies. These strategies are classified into preprocessing, in-processing, and post-processing methods, corresponding to the respective stages of AI model development, ensuring a structured approach to bias management throughout the lifecycle of AI models.[20,21]

## Results

This section outlines the major types of bias identified through our literature review, provides definitions and explanations for each, and then presents our detailed analyses of the studies included in the review.

### Types of bias

We identified 6 primary types of bias potentially present in the development of EHR-based AI models:

*Implicit bias:* Also known as prejudice bias, implicit bias occurs automatically and unintentionally.[22] It often stems from preexisting bias within the data (such as stereotypes and flawed case assumptions) which may occur in the data collection and data preprocessing steps. Utilizing biased data inevitably leads to biased outcomes. Racial bias, gender bias, and age bias fall under implicit bias.

*Selection bias*: This type of bias, also known as sampling bias or population bias, occurs when individuals, groups, or data used in analysis are not properly randomized during data preparation.[23,24] For instance, if an AI model is used to predict the mortality rate of patients with sepsis across the US but is only trained by data from a single hospital in a specific geographical area, it may not generalize well to a broader population, leading to skewed and inaccurate predictions.

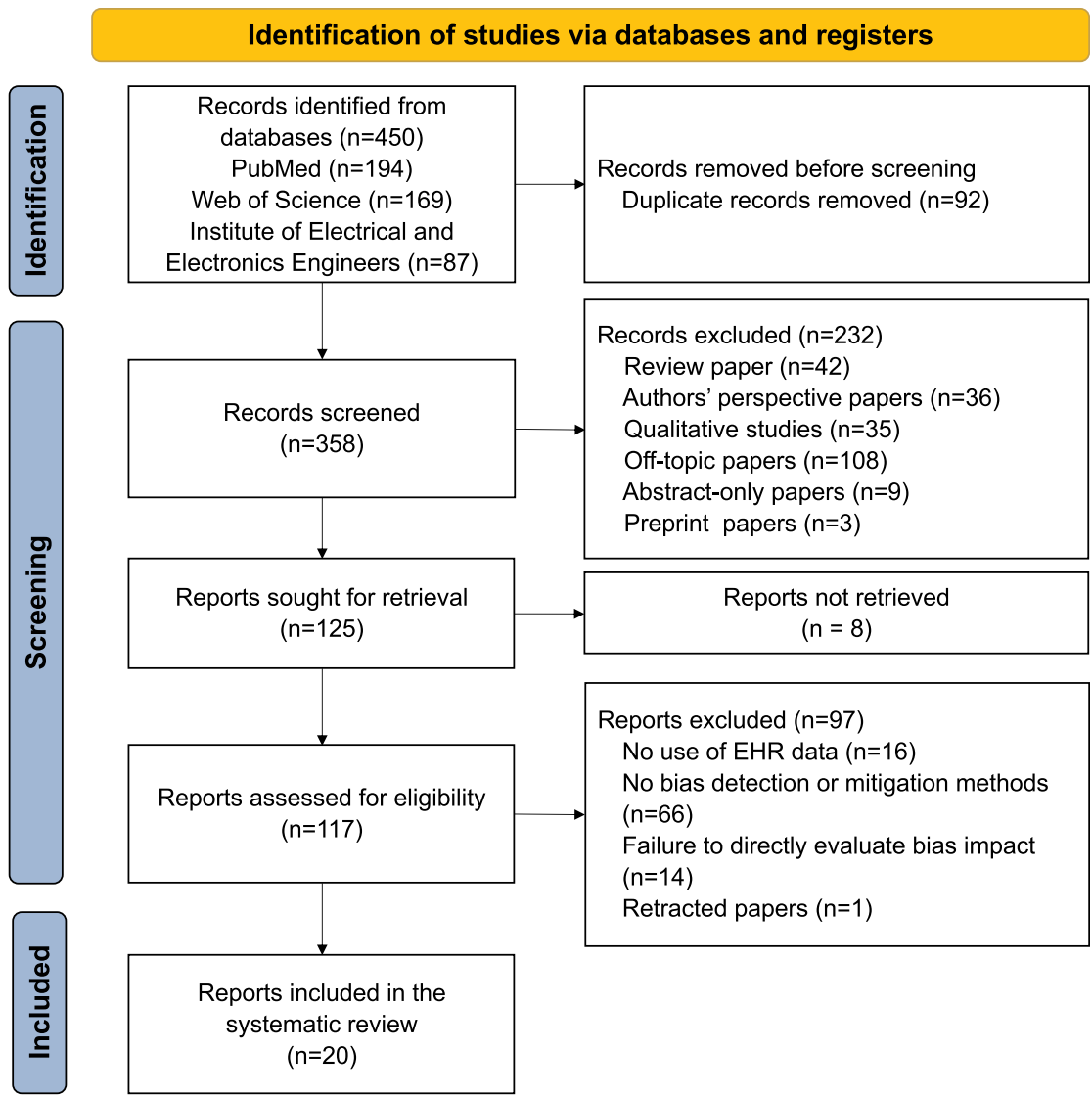

**Figure 1.** PRISMA flow diagram.

*Measurement bias*: This bias usually arises during the data collection stage of a study, often due to inaccuracies or incompleteness in data entries from clinicians or clinical devices.[25] Incorrect or biased labels resulting from coding errors or subjective interpretations by annotators can also impact the performance and validity of the machine learning models.

*Confounding bias*: Also known as association bias, confounding bias is a systematic distortion between an exposure and health outcome by extraneous factors.[24] Confounding bias could be introduced when collecting data as well as training models. For example, in a study predicting the readmission of patients, there could be confounding bias in the data related to socioeconomic status, as people with lower socioeconomic status may have limited access to healthcare resources making them more likely to have worse medical conditions. In this scenario, socioeconomic status is related to both the input medical conditions and the model predictions.

*Algorithmic bias*: This form of bias arises when the intrinsic properties of a model and/or its training algorithm create or amplify the bias in the training data. Algorithmic bias can create or amplify bias due to various factors, including imbalanced or misrepresentative training data, improper assumptions made by the model, lack of regulation in model processing, and so on.[7] For instance, linear regression models performing prediction on clinical data with complex features failing to meet the Gaussian distribution assumption might cause bias.

*Temporal bias*: This bias occurs when sociocultural prejudices and beliefs are systematically reflected,[26] especially when historical or longitudinal data is used to train models.[19] Such data likely embodies different healthcare practice patterns, outdated treatment and test records, different disease progression stages, and obsolete data recording processes that may negatively impact the performance on current data. Temporal bias could be introduced at any stage of AI application development.

### Article selection and screening results

Figure 1 outlines the article selection and screening process. We initially identified 450 articles, of which 92 were duplicates, leaving 358 articles. Upon abstract screening, 232 papers were excluded for not meeting our review criteria, including: review papers ($n = 42$), perspective articles ($n = 36$), off-topic articles ($n = 108$), qualitative studies ($n = 35$), abstract-only ($n = 9$), and preprint articles ($n = 3$). Additionally, 8 articles were excluded due to unavailability of the full text. 117 studies underwent full-text review, and 97 were further excluded, due to using non-EHR data ($n = 16$), lack of clear methods to detect or mitigate bias in AI models ($n = 66$), and failure to directly evaluate bias impact ($n = 14$). One paper was retracted before we finalized this review. Ultimately, 20 articles were included in the final analysis.

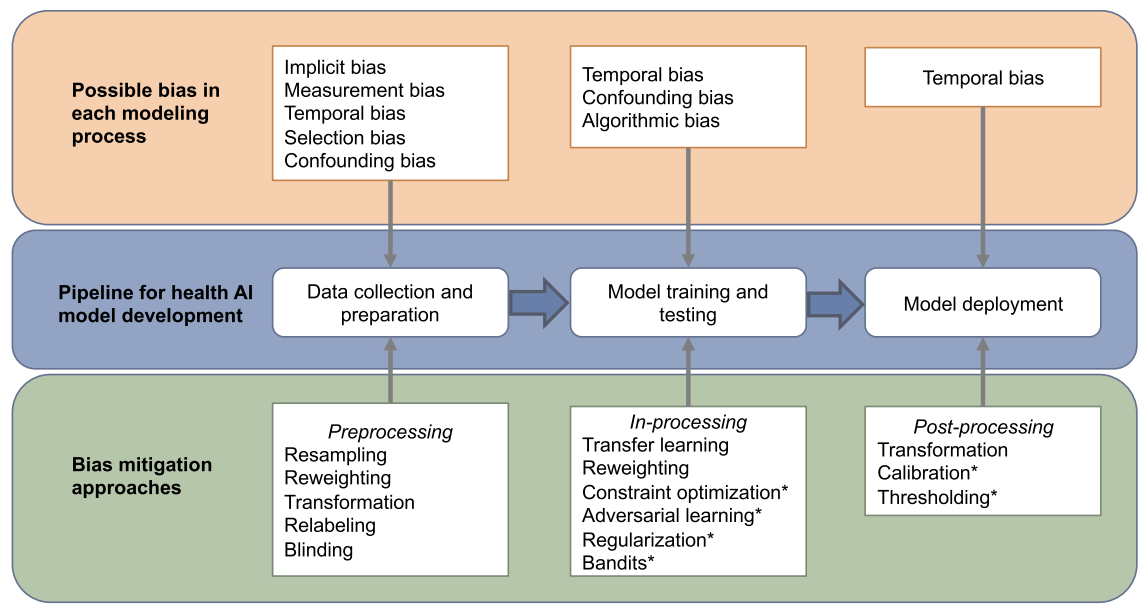

**Figure 2.** Bias handling workflow in artificial intelligence (AI) model development. The pipeline for the AI applications is shown in the blue box. Possible types of bias that could be introduced in each step are shown in the orange box, while possible handling approaches during each step of the model development are shown in green. Preprocessing handling methods refer to analyzing and adjusting the existing data set to preempt biases resulting from inadequate data during data collection and preparation. In-processing handling methods aim to handle bias during the model training and testing to avoid bias during training and eliminate bias from input data. Postprocessing handling methods account for handling model biases by interpreting or adjusting model outputs and correctly making use of the result. Approaches marked with * were approaches that were not covered by papers in this review.

### Research trends over time
The publication trends over the years are shown in Figure S1. All the final included articles were published after 2014. Notably, more than half of the papers (14 out of 20) were published in 2022 and 2023, indicating a growing interest in emphasizing bias and fairness when using EHR data in recent years.

### Tasks of research
To understand how bias might have influenced AI applications, we categorized the principal tasks of the AI models in each study, as detailed in Table 1. These EHR-based AI predictive tasks include diseases' diagnosis or risk prediction,[27–31] treatment effect or disease progression prediction,[4,32–35] mortality or survival prediction,[5,36–39] medication or test usage prediction, diseases' risk-association prediction,[40,41] health status classification,[42] and EHR missingness imputation.[43,44]

### Datasets
The datasets used in these studies were varied. Only 1 study utilized a commercial EHR dataset called Flatiron Health database.[43] The other studies used datasets derived from individual hospitals or integrated healthcare delivery systems. The size of the datasets varied drastically, with the largest dataset being over 1000 times larger than the smallest one. For example, Juhn's study enrolled a population of 555,[27] while Wolk's research was on a dataset of over 600 000 patients.[28] Five studies (25%) used 2 different EHR datasets for validation and evaluation.[29,31,37,41,44] Three studies published in 2023[37,41,44] indicate an increasing trend in addressing the generalizability and robustness of bias handling.

### Bias evaluation metrics
The studies included in this review employed a diverse array of metrics to assess bias, as detailed in Tables 2 and 3, with definitions described in Table 4. Eight studies (40%)[4,5,28,29,32,33,37,42] applied only performance metrics, such as sensitivity, specificity, accuracy, and area under receiver operating characteristic curve (AUROC), mean squared error (MSE). The remaining 12 (60%) employed fairness metrics and all of them focused on group fairness which tests for some form of statistical parity (eg, between positive outcomes, or errors) for members of different protected groups.[45] The classification of group-based fairness metrics[20] has 3 criteria: independence, separation, and sufficiency. Among the studies using fairness metrics, 3[30,34,40] applied parity-based metrics to measure the independence of the models from group information, which directly looks at how independent the predicted value among different groups. Nine[27,30,31,34–36,38–40] utilized confusion matrix-based metrics to measure the separation, which is the independence of the predicted value and groups conditional on true value. One used calibration-based metrics to measure the sufficiency, which is the independence of the true value and groups conditional for a given predicted value.[31] Two used

**Table 1.** Summary of the topics and datasets of those reviewed articles.

| Article | Task for AI model | Data sources | Sample size |
| --- | --- | --- | --- |
| Karlsson et al[32] | Prediction of adverse drug events | Stockholm Electronic Patient Record Corpus, Sweden | N/A |
| Hee et al[5] | Mortality prediction of 9 common diseases | MIMIC-III, MA, United States | 40 000 |
| Zhu et al[33] | Prediction of hospital re-admission rate | Over 10 South Florida regional hospitals, FL, United States | 92 827 |
| Huda et al[42] | Classification of the health condition or ability | Prospective pilgrims in Purworejo, Indonesia | 2425 |
| Allen et al[36] | Mortality prediction | MIMIC-III, MA, United States | 28 460 |
| Khoshnevisan et al[29] | Early detection of septic shock | Christiana Care Health System; Mayo Clinic, United States | 61 848; 74 463[a] |
| Juhn et al[27] | Risk prediction of asthma exacerbation | Mayo Clinic, United States | 555 |
| Wolk et al[28] | Risk prediction of influenza complications | Geisinger, PA, United States | 604 389 |
| Getz et al[43] | Missingness imputation of metastatic urothelial carcinoma patients' EHR | Flatiron Health database, United States | 3858 |
| Meng et al[38] | Mortality prediction for ICU admitted patients | MIMIC-IV, MA, United States | 43 005 |
| Wang et al[39] | Mortality prediction for sepsis patients | MIMIC-III, MA, United S | 11 719 |
| Röösli et al[31] | In-hospital mortality risk prediction | MIMIC-III; STARR, United States | 18 094; 6066 |
| Jiang et al[37] | Mortality prediction of respiratory system diseases | MIMIC-IV; eICU, United States | 11 167; 35 175 |
| Li et al[40] | Risk factors prediction of post-liver transplant | Organ Procurement and Transplantation Network, United States | 160 360 |
| Lee et al[4] | Disease progression prediction of glaucoma | Department of Ophthalmology in NYU Langone Health, NY, United States | 785 |
| Li et al[30] | Risk prediction of cardiovascular disease | Vanderbilt University Medical Center, TN, United States | 109 490 |
| Davoudi et al[34] | Pain level prediction of acute postoperative pain | University of Florida Health System/Shands Hospital, FL, United States | 14 263 |
| Cui et al[41] | Bipartite ranking for coronary heart disease | MIMIC-III; eICU, United States | 21 139; 17 402 |
| Yin et al[44] | Missingness imputation of temporal EHR data | MIMIC-III; eICU, United States | 13 112; 10 162 |
| Raza and Bashir[35] | ICU readmission prediction | MIMIC-III, MA, United States | 6500 |

[a] Khoshnevisan et al, Röösli et al, Jiang et al, Cui et al, and Yin et al had data from 2 dataset with different sample sizes.

score-based metrics to directly compare the predicted value.[41,43]

### Bias detection and mitigation characteristics

Of the 20 studies analyzed in this review, 11 (55%) pertained to implicit bias,[4,27,28,30–32,35,36,38–40] 6 (30%) pertained to selection bias,[5,33,40,42–44] 6 (30%) to algorithmic bias,[28,31,34,38,39,41] 1 to confounding bias,[29] 1 to measurement bias,[29] and 1 to temporal bias.[37] Six studies (30%) covered 2 types of bias,[28,29,31,38–40] while the rest focused on a single type of bias.

Of the 20 studies analyzed in this review, 5 (25%)[27,31,35,38,39] only detected the presence of bias in AI models by introducing quantitative measurement to identify and explain source of bias, as listed in Table 2. The other 15 (75%)[4,5,27–30,33,34,36,37,40–44] sought to mitigate biases. Among these mitigation approaches, 12 out of 15 (80%) reported improved performance after bias mitigation. In contrast, 2 studies (13.3%) observed that performance remained largely unchanged after bias mitigation,[34,43] and 1 (6.7%) found performance variability based on the evaluation metrics used.[32] Bias handling approaches and mitigation results are summarized in Table 3.

Figure 2 demonstrates the EHR-based AI application development stages, potential biases at each stage and corresponding bias mitigation approaches. When aligning the included articles with each stage, the details of each study's approach were described in Table 3. The bias mitigation methods of 11 studies (73.3%)[5,28–30,32,33,36,37,42–44] fell within the preprocessing step, using approaches including resampling, reweighting, transformation, relabeling, and blinding. Three studies (20%)[4,34,40] developed in-processing approaches including transfer learning and reweighting. Only 1 study (6.7%) applied a postprocessing bias mitigation method using transformation to identify bias from the model application.[41]

### Discussion

AI's growing role in healthcare has become particularly evident due to its unprecedented capability to harness EHR data for various tasks, such as mortality prediction, disease progression prediction, and risk assessment.[46] On the other hand, AI models may exhibit biases, and if left unaddressed, they have the potential to exacerbate healthcare disparities. There is an urgent need to thoroughly evaluate possible biases produced by AI models,[47] and assess the fairness of AI methods to reduce healthcare disparities.[48,49] In this systematic review, we have comprehensively identified and scrutinized potential biases in AI models developed from EHR data, revealing their potential influence on healthcare disparities. Our findings underline 6 major types of bias, with a notable focus on implicit and selection biases, and present an array of methods for their detection and mitigation. The study emphasizes the critical need for standardized, generalizable, and interpretable frameworks to ensure fairness in healthcare AI. By highlighting these issues, this review advocates for actionable strategies to promote equity in healthcare outcomes by improving the fairness of EHR-based AI models.

### Major types of bias in AI models using EHR data

In this review, all 6 types of bias have been covered by at least 1 study, with most focused on implicit and selection biases. In recent years, more studies have explored confounding and

Table 2. Paper summary for bias detection approaches.

| Article | Bias type | Evaluation metrics | Bias detection method | Bias detection summary |
| --- | --- | --- | --- | --- |
| Juhn et al[27] | Implicit bias | Accuracy, equal opportunity, equal odds, predictive equity | HOUSES is applied as a new feature, an individual-level SES measure based on 4 real property data variables of an individual housing unit after principal component factor analysis. | Asthmatic children with lower SES have larger balanced error rate than those with higher SES and has a higher proportion of missing information relevant to asthma care. |
| Röösli et al[31] | Implicit bias; algorithmic bias | Statistical parity, calibration-in-the-large | Statistical parity based on AUROC and AUPRC, and calibration are measured for 3 steps of model validation, respectively. | The predictive model is found hard to classify minority class instances correctly and fairly. |
| Wang et al[39] | Implicit bias; algorithmic bias | Statistical parity | The AUROC differences are compared between the entire cohort and those on the subpopulations by permutation tests. | Performance decreases for mortality prediction for minority racial and socioeconomic groups. |
| Meng et al[38] | Implicit bias; algorithmic bias | Statistical parity | AUROC for models and feature importance across different groups and the whole are compared. | Performance differences exist among different groups and models rely on different racial attributes across groups. |
| Raza et al[35] | Implicit bias | Equal opportunity, predictive equity, proportion parity, false negative rate parity | Confusion metrics are calculated and compared for each subgroup after training on the whole data. | Disparity based on 4 fairness evaluation, disparity among different socioeconomic groups is observed in the readmission prediction. |

Abbreviations: AUPRC = area under precision-recall (PR) curve, AUROC = area under the receiver operating characteristic curve, HOUSE index = HOUsing-based SocioEconomic Status measure, SES = socio-economic status.

algorithmic biases, which require a deeper understanding of data and models used in AI applications. Although confounding and algorithmic biases have been extensively studied in the fields of computer science and statistics,[50–52] their distinct impact on health disparities has not been thoroughly explored when applied to EHR data. This research gap underscores the need for a focused investigation into how these biases specifically impact health disparities within the context of EHR data analytics. While there is only 1 paper in this review[37] focused on temporal bias, other possible approaches to mitigate temporal bias might be feasible, including resampling data, transforming time-related features, and retraining the model or updating the learning rate during training to better learn the time-dependent data.[53,54] No study has yet developed a comprehensive pipeline to address more than 2 types of bias simultaneously. While previous research has concentrated on identifying and mitigating specific biases, future research may aim to tackle multiple biases concurrently throughout the model development process.

### Bias evaluation and fairness assessment

The bias evaluation methods in the reviewed studies highlight a methodological disparity, with approximately half employing fairness metrics to assess biases among groups. The rest relied on general performance metrics like accuracy, sensitivity, and specificity, which, while useful for evaluating overall model performance, may not detect subtle but significant disparities between groups.[55] This oversight can mask the differential impacts of AI models on diverse patient populations, inadvertently overlooking potential biases. Fairness metrics, designed to evaluate the equitable distribution of AI outcomes across demographic or clinical groups, are crucial for uncovering these disparities. As the field progresses, integrating fairness metrics with traditional performance metrics is vital to cultivate AI models that are both effective, and fair, ultimately supporting the goal of equitable healthcare through AI in EHR data analysis.

### The role of bias detection

Bias detection in this review primarily relied on analyzing performance differences among groups, with 1 study introducing the HOUSE index[27] as an innovative feature to better capture socioeconomic status and related biased features in the original data. These methods can effectively reveal implicit and algorithmic biases within AI models. Yet, their capacity to uncover hidden data patterns leading to bias remains limited. Two recent studies in 2023[56,57] that aimed to develop AI methods that directly detect and elucidate more nuanced bias in EHR data. Although these studies were not evaluated within AI models and hence were not included in this review, they signify a promising direction for the development of new technologies to detect more complex and nuanced biases in EHR data.

### Navigating bias mitigation approaches

Bias mitigation approaches in this review are categorized into 3 stages: preprocessing, in-processing, and postprocessing. Among these, preprocessing approaches are the most used approach to handle bias. These methods offer tangible, early-stage bias mitigation and might be readily adapted to clinical workflows. Resampling and reweighting, 2 major methods in this category, effectively address class or group imbalance by modifying the distribution of the training data. Other preprocessing strategies covered by this review include transformation to recover missing data, relabeling, and domain adaptation. Integrating multiple preprocessing approaches within a single pipeline to effectively address various bias types presents an opportunity for future research. However, they might be less effective in addressing feature correlations and may even lead to data loss. Thus, they are limited in mitigating confounding, algorithmic, and temporal bias.

**Table 3.** Paper summary for bias mitigation approaches by bias type.

| Article | Bias handling approach | Evaluation metrics | Bias research question | Bias handling method | Bias handling result |
|---|---|---|---|---|---|
| Implicit bias | | | | | |
| Allen et al[36] | *Preprocessing*: reweighting (reweight based on probability) | Equal opportunity | Racial bias exists between White and non-White racial groups in the early warning and mortality scoring systems. | Individual training examples are given weights based on mortality status and race within each age strata using a reweighting scheme. | The model reduces racial bias and improves accuracy comparing with existing mortality score predictors. |
| Karlsson et al[32] | *Preprocessing*: resampling | F1-score, AUROC, MSE | The data sparsity creates class bias and causes the prediction result to favor the majority class. | In the first resampling approach, $m$ new features are resampled a maximum of $n$ times if no informative feature is found, while in the second resampling approach 1 feature is resampled until an informative feature is found or until there are no more features. | Both approaches mitigate class bias. The suggested choice of approach to handle sparsity is highly dependent on the evaluation metrics. |
| Lee et al[4] | *In-processing*: transfer learning | MAE, MSE | Imbalanced ophthalmic clinical data arises healthcare disparities among racial groups. | Two-step transferred learning is used: first is transferring the information from source domain by fitting the linear model on the combined set and then correcting the bias by fitting the contrast solely on the target domain. | Compared with conventional approaches, transfer learning achieves better performance in MAE and MSE. |
| Li et al[30] | *Preprocessing*: resampling, blinding | Equal opportunity, statistical parity | Risk assessment predictive models can be biased because of systematic bias in the training data. | Three different approaches are used: removing protected attributes, resampling the imbalanced training dataset by sample size, and resampling by the proportion of people with CVD outcomes. | Removing protected attributes and resampling by sample size didn't significantly reduce the bias. Resampling by case proportion reduced the EOD and DI for gender groups but slightly reduced accuracy in many cases. |
| Selection bias | | | | | |
| Zhu et al[33] | *Preprocessing*: resampling (localize sampling) | AUROC | The population is highly skewed and biased for model because re-admission patients is a small proportion. | Localize sampling that uses LDA embedding assesses the locality of instances and allows the sampling process to bias to such instances. | Localized sampling helps to solve the sample imbalance issue for effective hospital re-admission prediction. |
| Huda[42] | *Preprocessing*: resampling (SMOTE + Neural Network) | Accuracy | Imbalanced datasets cause bias in Istitaah classification system. | SMOTE oversamples minority classes by creating a set of a new instance from minority class by interpolating several minority classes instances around the existing samples. | The combination of SMOTE and Neural Network gains the balanced dataset and was the most accurate in classification. |
| Hee et al[5] | *Preprocessing*: resampling (stratified random sampling) | AUROC, accuracy | Data quality assurance methods are needed to reduce bias when reusing clinical data for mortality prediction. | CDQA and MDQA identify most relevant variables to conduct stratified random sampling. | CDQA and MDQA stratify sampled inputs and improve predictive results in AUC and accuracy. |
| Getz et al[43] | *Preprocessing*: transformation (multiple imputation) | Percent bias | Missingness not at random may cause bias in machine learning model. | Missing values are imputed using multiple imputation using chain equation, random forest and denoising autoencoder, respectively. | Denoising autoencoders does not outperform the traditional multiple imputation methods. |

Table 3. (continued)

| Article | Bias handling approach | Evaluation metrics | Bias research question | Bias handling method | Bias handling result |
|---|---|---|---|---|---|
| Yin et al[44] | *Preprocessing*: transformation (data imputation) | AUPRC, MAE | Traditional models typically condition their model predictions on the partial observations and observational bias and degrade the performance. | The proposed model uses 3 subnetworks to impute missing data by propensity score adjustment. | The model outperforms current methods in binary data imputation, disease progression modeling, and mortality prediction. |
| Algorithmic bias | | | | | |
| Davoudi et al[34] | *In-processing*: reweighting | Equal opportunity, predictive equality, statistical parity | Predictive models systematically predict an outcome more likely for 1 group than another in some socioeconomic groups. | The weights of observations in each attribute-outcome combination in training the model are adjusted. | Reweighing the prediction models based on each protected attribute reduce the bias for some cases, but it introduced bias in some other cases where there was no bias. |
| Cui et al[41] | *Post-processing*: transformation (xOrder) | xAUC | Ranking models that rank positive instances higher than negative ones with poor fairness can cause systematic disparity across different protected groups. | The framework utilizes dynamic programming for adjusting ranking scores to optimize the ordering by minimizing an objective comprising a weighted sum of algorithm utility loss and ranking disparity. | The framework consistently achieves a better balance between the algorithm utility and ranking fairness on a variety of datasets with different metrics. |
| Temporal bias | | | | | |
| Jiang et al[37] | *Preprocessing*: transformation (time alignment) | AUROC, AUPRC, F1-score | Temporal bias can be introduced to longitudinal EHR data when treating patients with different disease progression state. | The model aligned patients' timeline to a shared timeline not simply treating hospital/ICU admission time as the "time-zero." | This time-alignment registration method enhances mortality prediction with at least a 1%-2% increase in evaluation metrics. |
| Implicit bias and selection bias | | | | | |
| Li et al[40] | *In-processing*: reweighting (dynamic reweighting) | AUROC, AUPRC, statistical parity, equal odds | Fairness discrepancy exists among sensitive attributes in predicting post-liver transplant risk factors. | Used fairness metrics including demographic disparity and equalize odds to measure disparity and optimize through iteration and reweight to balance the tasks in the loss function. | The algorithm reduced fairness disparity among all sensitive attributes and achieve task balance while maintaining accuracy. |
| Implicit bias and algorithmic bias | | | | | |
| Wolk et al[28] | *Preprocessing*: reweighting (inverse propensity score weighting) | AUROC | Gflu-CxFlag reveals significantly higher sensitivity for White than for Black individuals. | Inverse propensity weighting is used to correct for over-estimation on unvaccinated due to unrelated complications. | The reweighting improves the performance of the model and mitigate bias among different populations. |
| Confounding bias and measurement bias | | | | | |
| Khoshnevisan and Chi[29] | *Preprocessing*: relabeling (domain adaptation) | AUROC, Accuracy, sensitivity, specificity, F1-score | General scoring systems often lack the necessary sensitivity and specificity for identifying high risk septic shock patients due to covariate shift and systematic bias. | The model is a VRNN-based Adversarial Domain Separation construction, which separates 1 global-shared representation for all domains from local information. | The model outperforms state-of-the-art domain adaptation methods to address both covariate shift and systematic bias. |

Abbreviations: CDQA = contextual data quality assurance, CVD = cardiovascular disease, DI = disparity impact, EOD = equal opportunity difference, Gflu-CxFlag = Geisinger flu-complications flag, ICU = intensive care unit, LDA = latent Dirichlet allocation, MAE = mean absolute error, MDQA = mutual data quality assurance, MSE = mean squared error, SMOTE = synthetic minority oversampling technique, VRNN = variational recurrent neural network, xROC = CROSS-area under the cross-operating characteristic curve.

In-processing approaches offer dynamic strategies for bias mitigation during model training. Reweighting could dynamically reduce bias by changing the importance given to different training examples to focus more on the underrepresented classes or less on the overrepresented ones, helping to ensure that the model does not simply learn to predict the majority class or the bias present in the training data. Transfer learning, as proposed by Lee et al,[4] is particularly useful to mitigate bias with limited data, as this approach leverages models pre-trained on large datasets to enhance performance. Besides these, several other in-processing approaches, proposed for machine learning models, might also be feasible, including

**Table 4.** Evaluation metrics covered in reviewed articles.

| Term | Description |
| --- | --- |
| **Performance metrics** | |
| AUROC | Assessing the overall performance by measuring the area under the probability curve that plots the TPR against FPR at various threshold. |
| Accuracy | Measuring the correct proportion of predicted results. |
| F1-score | Measuring both the PPV and TPR. |
| Sensitivity | Same as the TPR, which measures the probability that the test assigns a diseased individual as positive. |
| Specificity | Same as the FNR of the model. |
| MAE | Measuring the average of the absolute difference between the predicted values and the actual values. |
| **Fairness metrics** | |
| Calibration-based metrics | |
| Calibration-in-the-large | Measuring the difference in average predicted and observed risk in each group. |
| Score-based metrics | |
| Percent bias | Measuring the average tendency of the predicted values to be larger or smaller than actual ones. |
| xAUC | A cross-group evaluation based on AUC that calculates how well the model distinguishes between positive examples in group A and negative examples in group B and vice versa. |
| Confusion Matrix-based metrics | |
| Equal odds | Measuring both TPR and FPR across groups, which is the percentage of actual negatives that are predicted as positive. |
| Equal opportunity | Representing the equal TPR across groups. |
| Predictive equality | Also known as predictive parity, which measures the difference of FPR across different groups. |
| Proportion parity | Measuring whether each group is represented proportionally to its share of the population. |
| False negative rate parity | Measuring the FNR across different groups. |
| Parity-based metrics | |
| Statistical parity | Also known as disparate impact or demographic parity that measures the difference in probabilities of a positive outcome across groups. |

Abbreviations: AUROC = area under the receiver operating characteristic curve, FNR = false negative rate, FPR = false positive rate, MAE = mean absolute value, MSE = mean squared error, NPV = negative predictive value, PPV = positive predictive value, TPR = true positive rate, xAUC = CROSS-area under the cross-operating characteristic curve.

constraint optimization to impose fairness constraints during the learning,[51] regularization to avoid overfitting,[21,58,59] and adversarial learning to determine the fairness of the training.[60–62]

In this review, only 1 study[41] employed postprocessing methods to adjust model output, an area currently underutilized. A previous survey[20] pointed out several other possible approaches to mitigate bias in machine learning models, including applying calibration to correct bias in the predicted probabilities,[63–65] and choosing proper thresholds to make decisions.[66–68] Postprocessing methods present a flexible and adaptable way to assess fairness, especially in black-box AI models. Future research should focus on robust postprocessing methods to effectively detect, mitigate, and clarify bias, which could greatly contribute to fairness in EHR-related AI applications.

## Research challenges and future directions

Addressing bias in AI models for healthcare purposes is a relatively recent research focus: there is a limited body of research directly dedicated to this crucial subject. A pressing need exists for establishing comprehensive guidelines for the use of evaluation metrics in AI models built from EHR data. These should include fairness evaluations to mitigate disparities among groups, alongside traditional performance metrics. Also, studies covered in this review all focused on group fairness.[69] Nevertheless, considering personalized healthcare using EHR data, future research efforts may focus more on developing and evaluation AI models which are fair to individuals.[45] None of these AI applications reviewed have been implemented in clinical settings. Further assessment is required to evaluate their practical use. Future research should focus on the following key areas to improve bias management in healthcare AI models:

1) *Data quality examination*: Ensuring the quality of EHR data is crucial for both clinical research and healthcare applications that might have a direct impact on patient care. Further research should continue to focus on the methods for improving the data quality like data cleaning, sampling and missingness imputation discussed in this review. This is crucial to reduce bias stemming from the quality of EHR data.

2) *Bias detection and mitigation pipeline*: Future research should attempt to develop a comprehensive pipeline that detects and mitigates multiple biases at each step of AI application development. Bias handling approaches should be integrated at each stage of the model development process, including data collection and preparation, model training and testing, model deployment, and result interpretation.

3) *Explainable bias for AI applications*: There is a need for more transparent AI models built from EHR data that can explain how and why biases occur. Approaches include statistics analysis, feature importance assessment, and ethical fairness considerations. Improving the interpretability will help to build fairer and more understandable clinical AI models for patients' care.

4) *Validation of AI model*: Validating AI models on diverse datasets is crucial to ensure their generalizability and robustness. This includes developing applications that are adaptable to a broader range of communities on multiple datasets and validating the influence of models in clinical environments.

## Limitations

First, despite comprehensive searches across 3 major databases, the limitations inherent in our keyword search strategy may have omitted potentially relevant papers. Second, the relatively modest number of papers included could limit the breadth of our analysis, which is likely due to the nascent stage of research within the scope specified by this review. Nonetheless, the detailed examination of each paper offers valuable insights into the current state of methodologies and their inherent constraints. Third, the interpretations and summarizations within this review are based on the authors' knowledge and understanding of the focused research area, introducing an element of subjectivity. Lastly, it is important to note that none of the models discussed in this review were tested in actual clinical settings, and as such, the impact of their bias handling methods on clinical outcomes has not yet been rigorously measured. The translation of these models into practice may present unforeseen challenges and learning opportunities that can only be identified through real-world application and evaluation.

## Conclusion

Bias in EHR-derived AI models has gained increasing attention by the research community. This review highlights current achievements and emphasizes the need for more in-depth research focused on developing standardized, generalizable and interpretable methods for detecting, mitigating and evaluating bias in AI models. As AI's presence in healthcare grows, it becomes critical to ensure that these technologies are equitable, thereby minimizing the risk of healthcare disparity due to biases. Continued efforts in this area are essential for improving the healthcare equity and maximizing the benefits of AI for healthcare.

## Author contributions

The following authors contributed to the following tasks: Li Zhou (Idea conception); Jiaqi Jiang, Liqin Wang, Feng Chen (development of database queries); Feng Chen, Jiaqi Jiang (article collection); Feng Chen, Jiaqi Jiang, Liqin Wang (article annotation); Feng Chen, Liqin Wang, Li Zhou (development of analysis methods); Feng Chen, Jiaqi Jiang (data analysis); Feng Chen, Jiaqi Jiang, Liqin Wang, Li Zhou (manuscript writing); and Feng Chen, Liqin Wang, Li Zhou (manuscript revision).

## Supplementary material

Supplementary material is available at *Journal of the American Medical Informatics Association* online.

## Funding

This work was supported by the National Library of Medicine under Grant No. 1R01LM014239.

## Conflicts of interest

None declared.

## Data availability

The data underlying this article are available in the article and in its online supplementary material.